\documentclass[letterpaper]{article} 
\usepackage[preprint]{aaai2027}  
\usepackage[hyphens]{url}  
\usepackage{graphicx} 
\usepackage{natbib}  
\usepackage{caption} 
\usepackage{subcaption}
\usepackage{algorithm}
\usepackage{algorithmic}

\usepackage{newfloat}
\usepackage{listings}
\DeclareCaptionStyle{ruled}{labelfont=normalfont,labelsep=colon,strut=off} 
\floatstyle{ruled}
\newfloat{listing}{tb}{lst}{}
\floatname{listing}{Listing}

\usepackage{booktabs}
\usepackage{algorithm}
\usepackage{algorithmic}
\usepackage{amsmath}
\usepackage{amssymb}
\usepackage{xcolor}
\usepackage{placeins}

\nocopyright

\newcommand{\method}{\textsc{DualAnchor}}
\newcommand{\task}{Sign Language Translation}

\newcommand{\SLT}{SLT}

\definecolor{deltaDown}{HTML}{0072B2}
\definecolor{deltaUp}{HTML}{D55E00}
\newcommand{\deltadown}[1]{\textcolor{deltaDown}{\ensuremath{\downarrow}\,\textbf{#1}}}
\newcommand{\deltaup}[1]{\textcolor{deltaUp}{\ensuremath{\uparrow}\,\textbf{#1}}}

\title{\textbf{DualAnchor}: Preserving Language Priors and Improving Lexical Fidelity in Gloss-Free Sign Language Translation}
\author{
    Hongbin Zhang\textsuperscript{\rm 1,\rm 2}\equalcontrib,
    Junhao Liu\textsuperscript{\rm 1}\equalcontrib,
    Xuefeng Bai\textsuperscript{\rm 1},
    Youcheng Pan\textsuperscript{\rm 2},
    Yang Xiang\textsuperscript{\rm 2},
    Kehai Chen\textsuperscript{\rm 1,\rm 2}
}
\affiliations{
    \textsuperscript{\rm 1}Institute of Computing and Intelligence, Harbin Institute of Technology, Shenzhen, China\\
    \textsuperscript{\rm 2}Peng Cheng Laboratory, Shenzhen, China\\
}

\begin{document}

\maketitle

\begin{abstract}
Recent advances in large language models (LLMs) have led sign language translation (SLT)—the task of converting sign-language videos into spoken-language text—to increasingly adopt LLMs as textual backbones.
However, despite LLMs' strong language modeling capabilities, existing LLM-based \SLT{} methods often undermine rather than exploit this language prior, producing disfluent translations---a failure we term \emph{language-prior degradation}.
Meanwhile, existing methods typically align videos and text at the sentence level; such alignment does not ensure accurate lexical details, creating a \emph{lexical fidelity gap}.
To address both issues, we propose \textbf{DualAnchor}, a gloss-free LLM-based \SLT{} training framework that couples two complementary anchors for linguistically fluent and visually faithful generation:
(i) \textbf{Token-level Prior Anchoring (TPA)} preserves the LLM's language prior by regularizing the multimodal decoder, at each decoding step, toward the next-token distribution of a frozen LLM conditioned on the same autoregressive prefix.
(ii) \textbf{Optimal Transport Alignment (OTA)} improves lexical fidelity by formulating visual--textual matching as entropy-regularized partial optimal transport, with Sinkhorn optimization inducing a soft alignment between visual tokens and textual content tokens under a cosine cost.
DualAnchor achieves strong overall performance on both PHOENIX-2014T and CSL-Daily.
Targeted analyses link the gains to TPA's fluency improvement and OTA's reduction of fine-grained lexical errors.
\end{abstract}

\section{Introduction}

\task{} (\SLT) translates sign-language videos into spoken-language text by recovering linguistic content from continuous visual signals and expressing it in target-language grammar~\cite{Camgoz_2018_CVPR,Camgoz_2020_CVPR,yin-read-2020-better}.
Recent advances in LLMs have led gloss-free \SLT{} methods to adopt a generation framework in which the model encodes the sign video, maps the resulting visual representation into the LLM's language space, and uses the pretrained LLM as the textual backbone for translation~\cite{Wong_2024_ICLR,Gong_2024_CVPR,chen-etal-2024-factorized,hwang-etal-2025-efficient}.
Within this framework, cross-modal alignment is usually learned at the sentence level through contrastive objectives over paired video and text representations or transferred semantic similarity~\cite{Zhou_2023_ICCV,Yin_2023_CVPR,jiang-etal-2024-signclip}.

\begin{figure}[t]
    \centering
    \includegraphics[width=1.0\columnwidth]{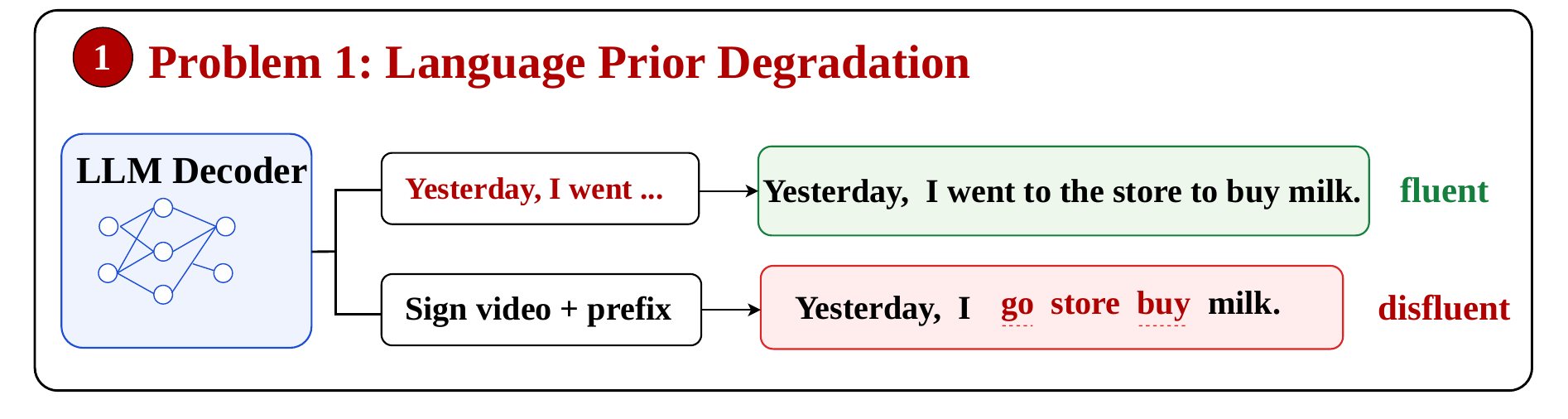}
    \includegraphics[width=1.0\columnwidth]{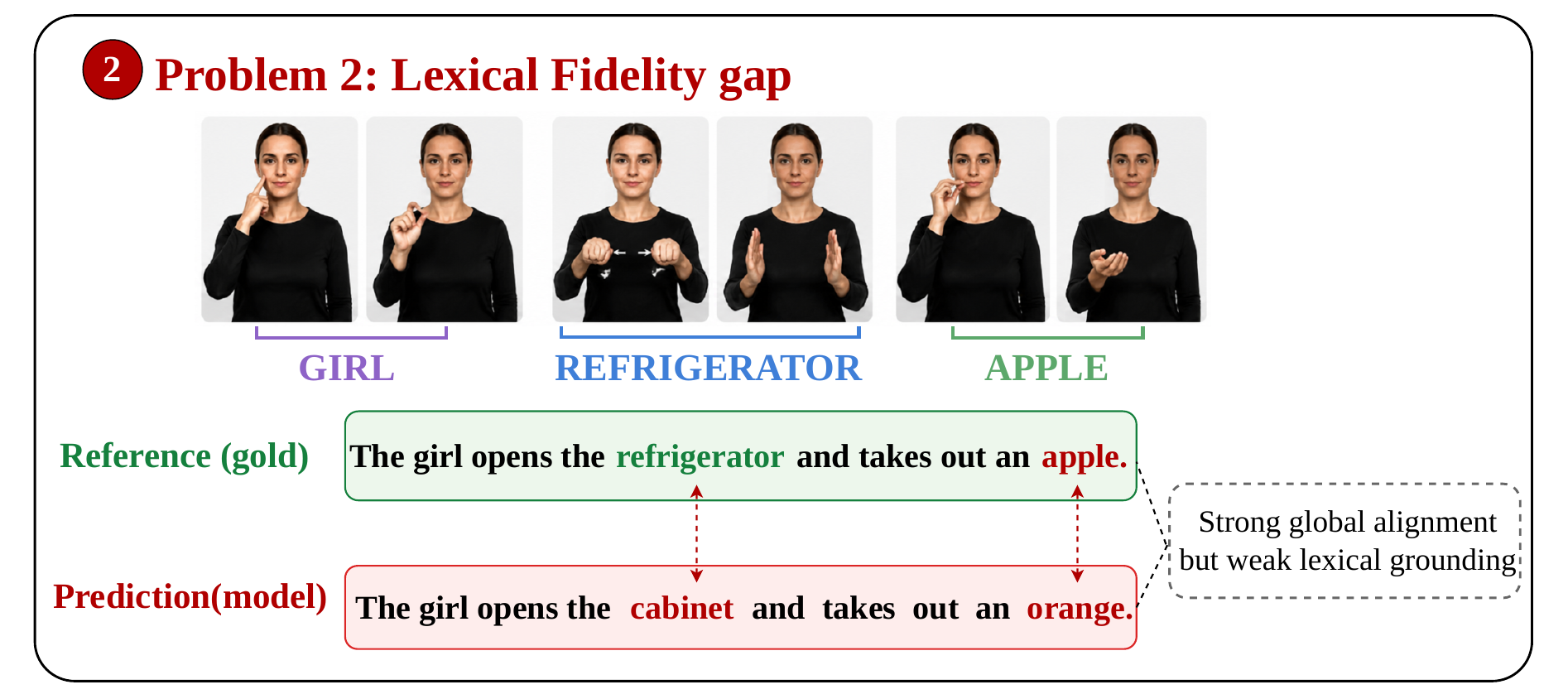}
    \caption{Motivating examples of the two issues in LLM-based \SLT{}.
    Top: \emph{language-prior degradation} produces disfluent text despite an LLM decoder.
    Bottom: strong global alignment can still coexist with weak lexical grounding, illustrating the \emph{lexical fidelity gap}.}
    \label{fig:motivation}
    \vspace{-15pt}
\end{figure}

However, this paradigm leaves two complementary issues unresolved.
Despite the strong language modeling capabilities of their LLM backbones, existing methods often undermine rather than exploit the pretrained language prior during visual adaptation.
Multimodal adaptation can shift the decoder's next-token distribution away from the linguistic structure acquired during text pretraining and reduce output fluency; we term this issue \emph{language-prior degradation}.
Sentence-level objectives align an entire sign video with its paired sentence, but they do not directly tie individual content words to local sign evidence.
A model can therefore achieve strong global video--text alignment while mistranslating actions, entities, or attributes; we call this discrepancy the \emph{lexical fidelity gap}.
The two issues affect complementary dimensions of translation quality: language-prior degradation weakens linguistic form, while the lexical fidelity gap compromises fine-grained content accuracy.
Figure~\ref{fig:motivation} provides an intuitive illustration.
The upper output is ungrammatical despite its LLM decoder.
The lower example shows strong global semantic alignment, yet replaces \emph{refrigerator} and \emph{apple} with visually unsupported \emph{cabinet} and \emph{orange}.

To address these issues, we propose \method{}, a gloss-free LLM-based \SLT{} framework with two anchors: token-level prior anchoring and optimal transport alignment.
\textbf{Token-level Prior Anchoring (TPA)} works in distribution space. Under the same autoregressive prefix, it regularizes the multimodal next-token distribution toward a frozen LLM prior and scales the constraint with the prior's confidence.
This design preserves pretrained decoding behavior while still letting visual evidence influence uncertain positions.
\textbf{Optimal Transport Alignment (OTA)} works in representation space. Entropy-regularized partial transport matches visual tokens to content tokens, while an unmatched sink absorbs unreliable correspondences.
The partial formulation handles visual--lexical granularity mismatch without forcing every token to align.
TPA preserves linguistic form, while OTA grounds lexical choice in fine-grained visual evidence.

We evaluate \method{} on PHOENIX-2014T~\cite{Camgoz_2018_CVPR} and CSL-Daily~\cite{Zhou_2021_CVPR}.
Across the two benchmarks, \method{} attains the best BLEU-4 among the compared gloss-free methods and the highest arithmetic mean across the ten reported metrics.
Targeted analyses associate the gains with more fluent generation and fewer fine-grained lexical errors, supporting the intended roles of TPA and OTA over generic model capacity.

Our contributions are threefold:
\begin{itemize}
    \item We identify two complementary LLM-based \SLT{} issues: \emph{language-prior degradation}, which hurts fluency during multimodal adaptation, and the \emph{lexical fidelity gap}, where sentence-level alignment still leaves content-word errors.
    \item We introduce \method{}, coupling confidence-aware token-level prior anchoring with entropy-regularized partial transport to preserve the pretrained language prior and ground lexical choices in visual evidence.
    \item On PHOENIX-2014T and CSL-Daily, \method{} achieves the best BLEU-4 among the compared gloss-free methods, and targeted analyses show better fluency and fewer fine-grained lexical errors.
\end{itemize}

\begin{figure*}[t]
    \centering
    \begin{minipage}[t]{0.322\textwidth}
        \centering
        \includegraphics[width=0.98\linewidth]{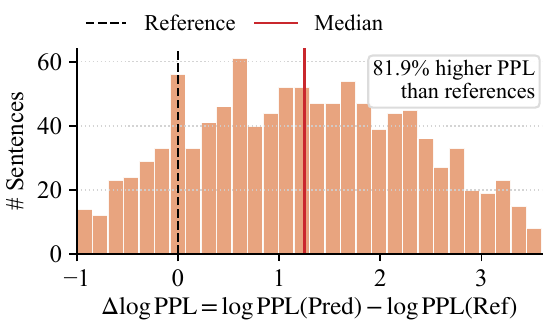}\\[-1pt]
        {\footnotesize\textbf{(a)} CSL-Daily: fluency gap}
    \end{minipage}\hfill
    \begin{minipage}[t]{0.322\textwidth}
        \centering
        \includegraphics[width=0.93\linewidth]{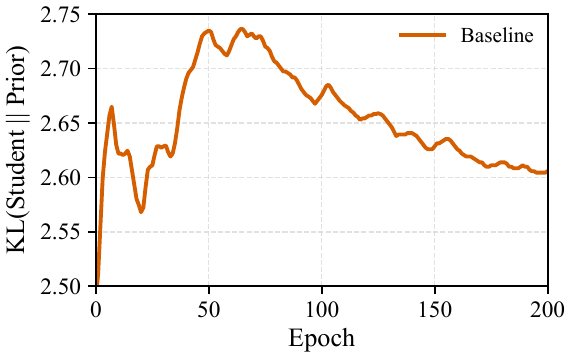}\\[-1pt]
        {\footnotesize\textbf{(b)} CSL-Daily: prior divergence}
    \end{minipage}\hfill
    \begin{minipage}[t]{0.322\textwidth}
        \centering
        \includegraphics[width=.9\linewidth]{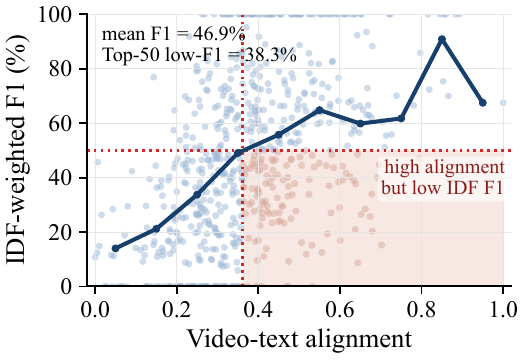}\\[-1pt]
        {\footnotesize\textbf{(c)} PHOENIX-2014T: global alignment vs. lexical fidelity}
    \end{minipage}
    \caption{Preliminary diagnostics of prior drift and lexical fidelity.
    (a) On CSL-Daily, $81.9\%$ of the adapted baseline's hypotheses have higher frozen-LM PPL than their paired references.
    (b) The student-to-prior KL divergence increases during multimodal adaptation and remains above its initial level.
    (c) On PHOENIX-2014T, $31.7\%$ of baseline hypotheses in the highest video--reference alignment quartile have IDF-weighted content-word F1 below $50\%$, showing that high global compatibility can coexist with lexical translation errors.}
    \vspace{-15pt}
    \label{fig:prelim-representative}
\end{figure*}

\section{Preliminary Analysis}
\label{sec:preliminary-analysis}

To characterize the two issues that motivate our method, we conduct two preliminary diagnostics.
We first examine whether multimodal adaptation is accompanied by an output-level fluency gap and increased divergence from the pretrained language prior.
We next ask whether high global video--reference compatibility can coexist with lexical errors.

\noindent\textbf{Diagnostic I: Does multimodal adaptation coincide with an output-level fluency gap and greater prior divergence?}
Given a sign-language video $V_i=\{v_{i,t}\}_{t=1}^{T_i}$ and its spoken-language reference $Y_i=\{y_{i,n}\}_{n=1}^{N_i}$, the controlled baseline generates $\widehat{Y}_i\sim p_{\theta}(Y\mid V_i)$ autoregressively.
We evaluate its hypotheses on the held-out sets of CSL-Daily (Chinese)~\cite{Zhou_2021_CVPR} and PHOENIX-2014T (German)~\cite{Camgoz_2018_CVPR}.
We measure fluency with a frozen external language model $q_{\phi}$, using Baichuan2-7B~\cite{Yang_2023_Baichuan2} for Chinese and Qwen2.5-7B~\cite{QwenTeam_2024_Qwen25} for German.
For a tokenized sentence $S=\{s_n\}_{n=1}^{|S|}$, we compute its perplexity and paired log-space fluency gap as
\begin{align}
    \operatorname{PPL}_{\phi}(S)
    &=\exp\!\left(
    -\frac{1}{|S|}\sum_{n=1}^{|S|}
    \log q_{\phi}(s_n\mid s_{<n})
    \right),
    \label{eq:prelim-ppl}\\
    \Delta_i^{\mathrm{flu}}
    &=\log \operatorname{PPL}_{\phi}(\widehat{Y}_i)
    -\log \operatorname{PPL}_{\phi}(Y_i).
    \label{eq:prelim-fluency-gap}
\end{align}
Lower PPL indicates higher target-language likelihood, while $\Delta_i^{\mathrm{flu}}>0$ means that the hypothesis is less probable than its paired reference under the same evaluator.

Figure~\ref{fig:prelim-representative}(a) reveals a pronounced output-level fluency gap on CSL-Daily: $81.9\%$ of the adapted baseline's hypotheses have higher PPL than their paired references, and the mean PPL differs by $3.4\times$ ($335.8$ versus $98.3$).

We next examine whether this output-level fluency gap co-occurs with increased divergence from the pretrained language prior by tracing the student-to-prior KL divergence throughout multimodal training.
At epoch $e$, let $p^s_{e,i,n}$ denote the baseline's next-token distribution and $p^p_{i,n}$ the frozen prior distribution under the same gold autoregressive prefix.
We average reverse KL over all valid target positions $\Omega$:
\begin{equation}
    D_e^{\mathrm{prior}}
    =\frac{1}{|\Omega|}
    \sum_{(i,n)\in\Omega}
    D_{\mathrm{KL}}\!\left(p^s_{e,i,n}\,\Vert\,p^p_{i,n}\right).
    \label{eq:prelim-prior-kl}
\end{equation}
Larger values indicate greater deviation from the frozen language prior under synchronized decoding contexts.

Figure~\ref{fig:prelim-representative}(b) shows that the KL divergence increases from $2.50$ to a peak of $2.77$ near epoch 65 and stabilizes around $2.63$ after epoch 220, remaining above its initial level.

\noindent\textbf{Finding I.}
Multimodal adaptation is associated with an output-level fluency gap and increased prior divergence: the adapted baseline's hypotheses receive substantially higher external-LM PPL than their paired references, while student-to-prior KL increases during training and remains above its initial level.
This coupled pattern motivates preserving the pretrained language prior during visual adaptation.

\noindent\textbf{Diagnostic II: Can high global video--reference compatibility coexist with lexical translation errors?}
Using the same baseline outputs and held-out samples as in Diagnostic I, we examine whether gold video--reference pairs that are well matched at the sentence level can still have corresponding hypotheses with weak content-word fidelity.
Let $g_v$ and $g_t$ denote the video and text encoders used by the sentence-level retrieval diagnostic.
For each gold video--reference pair, we measure global compatibility as $s_i^{\mathrm{sent}}=\operatorname{sim}\!\left(g_v(V_i),g_t(Y_i)\right)$, where larger values indicate stronger sentence-level compatibility between the video and its reference translation in the diagnostic representation space.
To measure fine-grained lexical fidelity, let $K_i^{h}=K(\widehat{Y}_i)$ and $K_i^{r}=K(Y_i)$ denote the normalized content-word sets of the hypothesis and reference, respectively, and define $W(A)=\sum_{w\in A}\operatorname{idf}(w)$.
The IDF-weighted content-word F1 is
\begin{equation}
    F_{1,i}^{\mathrm{idf}}
    =\frac{2W(K_i^{h}\cap K_i^{r})}
    {W(K_i^{h})+W(K_i^{r})}.
    \label{eq:prelim-idf-f1}
\end{equation}
The resulting score measures content-word agreement between the baseline hypothesis and its reference.
IDF assigns greater weight to rare and informative words, while F1 captures both missing reference content and unmatched content introduced by the hypothesis.
Content-word-aware translation likewise treats lexical units as unequally important to sentence meaning~\cite{chen-etal-2020-content}.

Figure~\ref{fig:prelim-representative}(c) shows that lexical errors remain common even among globally well-matched gold video--reference pairs: within the highest alignment quartile, $31.7\%$ of baseline hypotheses have IDF-weighted content-word F1 below $50\%$.
Thus, fine-grained lexical errors can persist despite high global video--reference compatibility.

\noindent\textbf{Finding II.}
High global video--reference compatibility can coexist with substantial lexical translation errors: even within the highest alignment quartile, $31.7\%$ of baseline hypotheses have IDF-weighted content-word F1 below $50\%$.
This coexistence motivates visual--textual alignment at a finer lexical granularity than sentence-level matching.



\section{Methodology}
\label{sec:methodology}

\subsection{Problem Formulation}

\begin{figure*}[t]
    \centering
    \includegraphics[width=\textwidth]{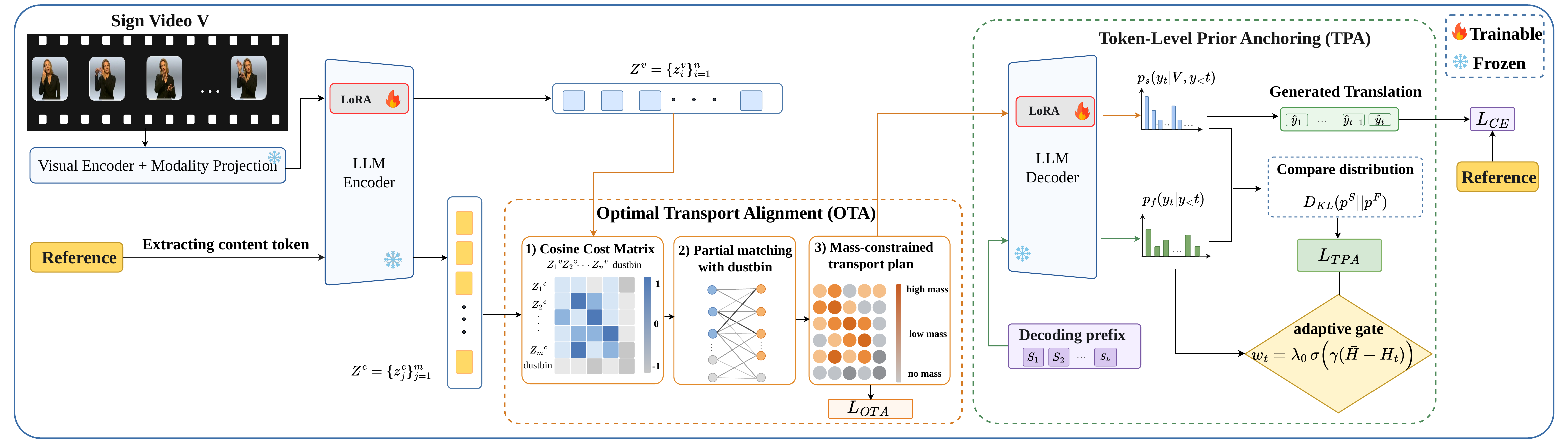}
    \vspace{-10pt}
    \caption{Overview of \method{}.
    TPA anchors the student next-token distribution to a frozen language prior under the same autoregressive prefix, while OTA learns partial correspondences between visual tokens and textual content tokens.
    Both modules operate only during training.}
    \vspace{-10pt}
    \label{fig:method-overview}
\end{figure*}

Given a sign video $V=\{v_\ell\}_{\ell=1}^{T}$ and its spoken-language translation $Y=\{y_t\}_{t=1}^{N}$, an LLM-based \SLT{} model maps $V$ to decoder-compatible visual tokens,
\begin{equation}
    \mathbf{Z}^{v}=F_{\psi}(V)
    =\{\mathbf{z}^{v}_{i}\}_{i=1}^{n},
    \qquad \mathbf{z}^{v}_{i}\in\mathbb{R}^{d},
    \label{eq:visual-tokens}
\end{equation}
where $F_{\psi}$ comprises the backbone's visual encoder and modality projection.
The multimodal student factorizes the translation probability autoregressively as
\begin{equation}
    p_{\theta_s}(Y\mid V)
    =\prod_{t=1}^{N}p_{\theta_s}
    \!\left(y_t\mid \mathbf{Z}^{v},y_{<t}\right),
    \label{eq:slt-factorization}
\end{equation}
and learns from the teacher-forced translation loss
\begin{equation}
    \mathcal{L}_{\mathrm{CE}}
    =-\frac{1}{|\Omega|}
    \sum_{(b,t)\in\Omega}
    \log p_{\theta_s}
    \!\left(y_{b,t}\mid \mathbf{Z}^{v}_{b},y_{b,<t}\right),
    \label{eq:ce-loss}
\end{equation}
where $\Omega$ contains the non-padding target positions in a minibatch.
Cross-entropy supervises the observed target but does not explicitly preserve the pretrained trajectory of next-token distributions or impose local geometry between visual evidence and lexical units.
\method{} therefore adds two complementary training constraints (Figure~\ref{fig:method-overview}): TPA anchors linguistic form in distribution space, and OTA anchors lexical content in representation space.
Together, TPA and OTA target a translation's linguistic form and visual content.

\subsection{Token-Level Prior Anchoring}
\label{sec:tpa}

Knowledge distillation transfers teacher behavior to sequence models and multilingual translation systems~\cite{kim-rush-2016-sequence,sun-etal-2020-knowledge-distillation}. TPA adapts this principle to token-level preservation during multimodal training.

\noindent\textbf{Prefix-synchronized language prior.}
TPA retains a frozen copy of the language model, $p_{\theta_p}$, to preserve the generative structure acquired during pretraining.
Let $s_b$ be a context prompt and $\pi_{b,t}=[s_b;y_{b,<t}]$ the context-augmented gold autoregressive prefix.
The student and prior share $\pi_{b,t}$ and the vocabulary; only the student observes the sign video:
\begin{align}
    p^{s}_{b,t}(k)
    &=p_{\theta_s}
      \!\left(y_t=k\mid \mathbf{Z}^{v}_{b},\pi_{b,t}\right),
      \label{eq:student-distribution}\\
    p^{p}_{b,t}(k)
    &=p_{\theta_p}
      \!\left(y_t=k\mid \pi_{b,t}\right).
      \label{eq:prior-distribution}
\end{align}
\emph{English Context} prepends an English task instruction; \emph{Target Context}, used by default, prepends its target-language rendering.
In both settings, $s_b$ is fixed before decoding and the target-dependent part of $\pi_{b,t}$ is exactly $y_{b,<t}$; the prefix-only variant sets $s_b=\varnothing$.
We freeze $\theta_p$ and detach its probabilities from gradient computation.
The synchronized prefix isolates multimodal adaptation's effect at each next-token position.

\noindent\textbf{Confidence-adaptive distribution anchoring.}
The frozen prior is not equally reliable at every position: strict imitation at uncertain positions can suppress useful visual evidence, whereas a sharp prior provides a stronger linguistic constraint.
TPA measures this reliability with token entropy
\begin{equation}
\begin{aligned}
    H_{b,t}
    &=-\sum_{k\in\mathcal{V}}
    p^{p}_{b,t}(k)\log p^{p}_{b,t}(k),\\
    \bar H
    &=\frac{1}{|\Omega|}\sum_{(b,t)\in\Omega}H_{b,t},
\end{aligned}
    \label{eq:prior-entropy}
\end{equation}
where $\mathcal{V}$ is the target vocabulary.
TPA then applies the confidence gate $\omega_{b,t}=\lambda_{0}\sigma\!\left(\gamma(\bar H-H_{b,t})\right)$, where $\lambda_0$ sets the maximum anchoring scale, $\gamma$ controls gate sharpness, and $\sigma(\cdot)$ is the sigmoid function.
Low-entropy positions receive stronger prior guidance; high-entropy positions leave the student more freedom to follow the video.

TPA averages reverse KL over valid target positions:
\begin{equation}
    \mathcal{L}_{\mathrm{TPA}}
    =\frac{1}{|\Omega|}
    \sum_{(b,t)\in\Omega}
    \omega_{b,t}
    D_{\mathrm{KL}}
    \!\left(p^{s}_{b,t}\,\|\,p^{p}_{b,t}\right).
    \label{eq:tpa-loss}
\end{equation}
This direction penalizes probability mass that the adapted student assigns to tokens unsupported by the frozen prior.
TPA regularizes the predictive distribution rather than prescribing a decoded sentence, preserving linguistic structure while retaining visual conditioning.

\subsection{Optimal Transport Alignment}
\label{sec:ota}

\noindent\textbf{Content-focused cross-modal geometry.}
Sentence-level alignment captures global video-text correspondence but leaves individual actions, entities, attributes, and temporal expressions ungrounded.
OTA addresses this gap by aligning the visual tokens in Equation~\eqref{eq:visual-tokens} with content-token representations $\mathbf{Z}^{c}=\{\mathbf{z}^{c}_{j}\}_{j=1}^{m}$, where $\mathbf{z}^{c}_{j}\in\mathbb{R}^{d}$.
The content selector excludes padding, special symbols, punctuation, and weakly semantic function tokens so that non-lexical units do not dominate the alignment.
After $\ell_2$ normalization, the pairwise cosine cost is
\begin{equation}
    D_{ij}
    =1-
    \frac{(\mathbf{z}^{v}_{i})^{\top}\mathbf{z}^{c}_{j}}
    {\|\mathbf{z}^{v}_{i}\|_2\,\|\mathbf{z}^{c}_{j}\|_2},
    \qquad
    \mathbf{D}\in\mathbb{R}^{n\times m}.
    \label{eq:ot-cost}
\end{equation}
The soft cost retains plausible correspondences because sign duration and text tokenization differ in granularity.

\noindent\textbf{Entropy-regularized partial transport.}
Entropy regularization makes optimal transport differentiable and efficiently solvable by Sinkhorn scaling~\cite{NIPS2013_af21d0c9}, while partial transport permits unmatched mass when the two supports contain unreliable correspondences~\cite{chapel-etal-2020-partial}.
Let $\mathbf{a}=\frac{1}{n}\mathbf{1}_{n}$ and $\mathbf{b}=\frac{1}{m}\mathbf{1}_{m}$ denote uniform visual and textual masses, and let $\rho\in(0,1]$ be the mass reserved for genuine cross-modal matching.
We define the partial transport polytope
\begin{equation}
\begin{aligned}
    \Pi_{\rho}(\mathbf{a},\mathbf{b})
    =\big\{\mathbf{P}\in\mathbb{R}_{+}^{n\times m}:~
    &\mathbf{P}\mathbf{1}_{m}\leq\mathbf{a},~
    \mathbf{P}^{\top}\mathbf{1}_{n}\leq\mathbf{b},\\
    &\mathbf{1}_{n}^{\top}\mathbf{P}\mathbf{1}_{m}=\rho\big\}.
\end{aligned}
    \label{eq:partial-polytope}
\end{equation}
Introduce slack masses $\mathbf{s}^{v}=\mathbf{a}-\mathbf{P}\mathbf{1}_{m}$ and
$\mathbf{s}^{c}=\mathbf{b}-\mathbf{P}^{\top}\mathbf{1}_{n}$, and set
$\bar{\mathbf P}=\bigl[\begin{smallmatrix}\mathbf P&\mathbf s^{v}\\(\mathbf s^{c})^{\top}&0\end{smallmatrix}\bigr]$,
$\bar{\mathbf a}=[\mathbf a;1-\rho]$, and $\bar{\mathbf b}=[\mathbf b;1-\rho]$.
Let $\bar{\Pi}_{\rho}(\bar{\mathbf a},\bar{\mathbf b})$ contain balanced couplings with these marginals and $\bar P_{n+1,m+1}=0$.
With $\bar{\mathbf D}$ equal to $\mathbf D$ on the real block and constant on dustbin edges, we solve
\begin{equation}
    \bar{\mathbf{P}}^{\star}
    =\arg\min_{\bar{\mathbf{P}}\in
    \bar{\Pi}_{\rho}(\bar{\mathbf a},\bar{\mathbf b})}
    \langle\bar{\mathbf{P}},\bar{\mathbf{D}}\rangle
    -\tau\,\mathcal{H}(\bar{\mathbf{P}}),
    \label{eq:partial-ot}
\end{equation}
where $\tau>0$ controls transport smoothness and
$\mathcal{H}(\bar{\mathbf{P}})=-\sum_{i,j}\bar P_{ij}(\log \bar P_{ij}-1)$, with $0\log 0=0$.
Thus $\mathcal{H}(\bar{\mathbf P})=\mathcal{H}(\mathbf P)+\mathcal{H}(\mathbf s^{v})+\mathcal{H}(\mathbf s^{c})$, explicitly regularizing unmatched capacity, while the real block transports exactly $\rho$ mass.
Let $\mathbf M$ be one except for $M_{n+1,m+1}=0$; masked Sinkhorn scaling gives
\begin{equation}
    \mathbf{K}=\mathbf M\odot\exp(-\bar{\mathbf D}/\tau),
    \qquad
    \bar{\mathbf P}^{\star}
    =\operatorname{diag}(\mathbf{u})\,
     \mathbf{K}\,
     \operatorname{diag}(\mathbf{r}),
    \label{eq:sinkhorn-plan}
\end{equation}
where Sinkhorn iterations alternately rescale $\mathbf u$ and $\mathbf r$ to match $\bar{\mathbf a}$ and $\bar{\mathbf b}$.
The real block of $\bar{\mathbf P}^{\star}$ gives $\mathbf P^{\star}$, while the dustbins absorb the remaining $1-\rho$ mass.
OTA then minimizes the average cost carried by genuine matches:
\begin{equation}
    \mathcal{L}_{\mathrm{OTA}}
    =\frac{\langle\mathbf{P}^{\star},\mathbf{D}\rangle}
    {\sum_{i=1}^{n}\sum_{j=1}^{m}P^{\star}_{ij}}.
    \label{eq:ota-loss}
\end{equation}
The transport plan supplies token-level structure, while the dustbin prevents ambiguous or semantically empty units from creating spurious supervision.

\subsection{Joint Optimization and Inference}
\label{sec:joint-training}

The two anchors regularize complementary model objects and are combined with translation supervision as
\begin{equation}
    \mathcal{L}
    =\mathcal{L}_{\mathrm{CE}}
    +\alpha_{\mathrm{TPA}}\mathcal{L}_{\mathrm{TPA}}
    +\alpha_{\mathrm{OTA}}\mathcal{L}_{\mathrm{OTA}},
    \label{eq:full-objective}
\end{equation}
where $\alpha_{\mathrm{TPA}}$ and $\alpha_{\mathrm{OTA}}$ balance linguistic stability and visual grounding.
At inference, the frozen prior, content selector, and Sinkhorn solver are removed, and the student follows the original autoregressive decoding path.
\method{} therefore changes training supervision without adding test-time teacher access or optimal-transport computation.

\section{Experiments}
\label{sec:experiments}

\begin{table}[t]
\centering
\footnotesize
\setlength{\tabcolsep}{3.4pt}
\renewcommand{\arraystretch}{0.94}

{\small\textbf{PHOENIX-2014T}\par}
\begin{tabular}{@{}lrrrrr@{}}
\toprule
Method & B1 & B2 & B3 & B4 & R-L \\
\midrule
GFSLT-VLP & 43.71 & 33.18 & 26.11 & 21.44 & 42.49 \\
FLa-LLM & 46.29 & 35.33 & 28.03 & 23.09 & 45.27 \\
Sign2GPT & 49.54 & 35.96 & 28.83 & 22.52 & 48.90 \\
SignLLM & 45.21 & 34.78 & 28.05 & 23.40 & 44.49 \\
BeyondGloss & \underline{52.38} & \underline{38.57} & 30.74 & 25.49 & \textbf{52.89} \\
MMSLT & 48.92 & 38.12 & 30.79 & 25.73 & 47.97 \\
SpaMo & 49.80 & 37.32 & 29.50 & 24.32 & 46.57 \\
SCL-SLT & 48.72 & 38.19 & \underline{31.04} & \underline{26.00} & 47.02 \\
\addlinespace[1pt]
\textbf{\method{} (ours)} & \textbf{53.93} & \textbf{41.28} & \textbf{33.18} & \textbf{27.60} & \underline{50.62} \\
\bottomrule
\end{tabular}

\vspace{3pt}
{\small\textbf{CSL-Daily}\par}
\begin{tabular}{@{}lrrrrr@{}}
\toprule
Method & B1 & B2 & B3 & B4 & R-L \\
\midrule
GFSLT-VLP & 39.37 & 24.93 & 16.26 & 11.00 & 36.44 \\
FLa-LLM & 37.13 & 25.12 & 18.38 & 14.20 & 37.25 \\
Sign2GPT & 41.75 & 28.73 & 20.60 & 15.40 & 42.36 \\
SignLLM & 39.55 & 28.13 & 20.07 & 15.75 & 39.91 \\
BeyondGloss & \underline{53.12} & 38.63 & 27.82 & 21.53 & \textbf{53.46} \\
MMSLT & 49.87 & 36.37 & 27.29 & 21.11 & 48.92 \\
SpaMo & 48.90 & 36.90 & 26.78 & 20.55 & 47.46 \\
SCL-SLT & 52.81 & \underline{39.28} & \underline{29.82} & \underline{23.25} & \underline{51.08} \\
\addlinespace[1pt]
\textbf{\method{} (ours)} & \textbf{53.15} & \textbf{39.84} & \textbf{30.65} & \textbf{24.21} & 49.46 \\
\bottomrule
\end{tabular}
\caption{Gloss-free SLT results on the official test sets. B$n$ and R-L denote BLEU-$n$ and ROUGE-L; \textbf{bold} and \underline{underlining} mark the best and second-best scores.}
\label{tab:main-results}
\vspace{-10pt}
\end{table}


\noindent\textbf{Datasets and metrics.}
We evaluate on PHOENIX-2014T and CSL-Daily.
PHOENIX-2014T contains 8,257 German Sign Language weather videos with German translations and uses the standard 7,096/519/642 train/development/test split~\cite{Camgoz_2018_CVPR}; CSL-Daily contains 20,654 Chinese Sign Language videos on daily-life topics with Chinese translations and uses the standard 18,401/1,077/1,176 split~\cite{Zhou_2021_CVPR}.
We report BLEU-1--BLEU-4~\cite{papineni-etal-2002-bleu} and ROUGE-L~\cite{lin-2004-rouge}.

\noindent\textbf{Baselines.}
We compare eight published gloss-free systems spanning visual-language pretraining (GFSLT-VLP~\cite{Zhou_2023_ICCV}), LLM adaptation (FLa-LLM~\cite{chen-etal-2024-factorized}, Sign2GPT~\cite{Wong_2024_ICLR}, SignLLM~\cite{Gong_2024_CVPR}, and SpaMo~\cite{hwang-etal-2025-efficient}), enriched visual or MLLM supervision (BeyondGloss~\cite{Asasi_2025_BMVC} and MMSLT~\cite{Kim_2025_ICCV}), and contrastive alignment (SCL-SLT~\cite{lai-etal-2026-selective}).

\noindent\textbf{Implementation details.}
We fine-tune the language backbone with LoRA~\cite{hu2022lora} using AdamW~\cite{loshchilov2019decoupled}. Unless otherwise specified, experiments use 4 GPUs, at most 500 epochs, and an early-stopping patience of 50 validation rounds. We use a learning rate of $3\times10^{-4}$ with cosine decay, select checkpoints by the best validation BLEU-4, and decode with beam size 6 and length penalty 1.0; the auxiliary objectives are training-only.

\noindent\textbf{Main Results.}
\method{} outperforms SpaMo on all ten metrics and gives the strongest higher-order BLEU results on both datasets (Table~\ref{tab:main-results}).
BLEU-4 rises from 24.32 to 27.60 on PHOENIX-2014T and from 20.55 to 24.21 on CSL-Daily, while the remaining BLEU and ROUGE-L scores improve consistently.
Across all compared methods, \method{} achieves the highest arithmetic mean over the ten reported metrics.
Gains span sign languages and domains and are strongest in multi-token correspondence.

\begin{table}[htbp]
\centering
\small
\setlength{\tabcolsep}{7pt}
\begin{tabular}{@{}llrr@{}}
\toprule
Backbone & Method & B4 & R-L \\
\midrule
NLLB & SpaMo & 17.89 & 43.29 \\
     & MMSLT & 17.96 & 41.17 \\
     & \textbf{\method{} (ours)} & \textbf{24.21} & \textbf{49.46} \\
\addlinespace[2pt]
mBART & SpaMo & 19.70 & 46.17 \\
      & MMSLT & 20.21 & \textbf{50.58} \\
      & \textbf{\method{} (ours)} & \textbf{22.35} & 49.75 \\
\addlinespace[2pt]
mT0 & SpaMo & 19.59 & 42.12 \\
    & MMSLT & 19.92 & 43.32 \\
    & \textbf{\method{} (ours)} & \textbf{22.56} & \textbf{47.75} \\
\bottomrule
\end{tabular}
\caption{CSL-Daily cross-backbone results. \textbf{Bold} marks the best completed score for each backbone. Rows share cached visual features, decoding settings, and the evaluation script.}
\label{tab:cross-backbone}
\vspace{-3pt}
\end{table}

\noindent\textbf{Cross-backbone generalization.}
Table~\ref{tab:cross-backbone} evaluates TPA and OTA with NLLB~\cite{nllbteam2024scaling}, mBART~\cite{liu-etal-2020-multilingual-denoising}, and mT0~\cite{muennighoff-etal-2023-crosslingual} under the same protocol.
Across completed runs, \method{} leads BLEU-4 on all three backbones and improves both metrics over SpaMo in every matched comparison.
The gains are largest with NLLB and remain positive on the other two backbones.
On mBART, MMSLT leads ROUGE-L but \method{} leads BLEU-4, making the ranking metric-dependent.

\section{Analysis}
\label{sec:analysis}
\flushbottom
\begingroup
This section asks four questions: (i) \textbf{RQ1:} How does TPA affect fluency and linguistic structure, and do these effects align with prior preservation? (ii) \textbf{RQ2:} Does OTA improve fine-grained lexical grounding and use local visual evidence as intended? (iii) \textbf{RQ3:} How sensitive are TPA and OTA to their loss weights? (iv) \textbf{RQ4:} Which TPA and OTA design choices govern translation quality and module behavior?

\smallskip
\noindent\textbf{RQ1: How does TPA affect fluency and linguistic structure, and do these effects align with prior preservation?}
Using SpaMo on CSL-Daily, we compare otherwise identical models with and without TPA to test how prior preservation relates to fluency and linguistic form.

\noindent\textbf{PPL and KL dynamics.}
At matched checkpoints, lower frozen-Baichuan2-7B corpus PPL indicates better external fluency, while lower mean student--prior KL under shared gold prefixes indicates less drift from the frozen prior.

\begin{figure}[htbp]
    \centering
    \begin{subfigure}[t]{0.50\linewidth}
        \centering
        \includegraphics[width=\linewidth]{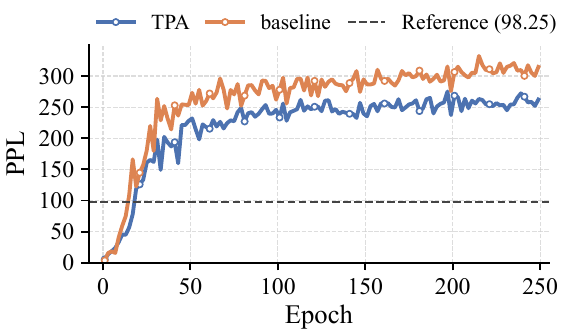}
        \caption{External PPL}\label{fig:tpa-external-ppl}
    \end{subfigure}\hfill
    \begin{subfigure}[t]{0.47\linewidth}
        \centering
        \includegraphics[width=\linewidth]{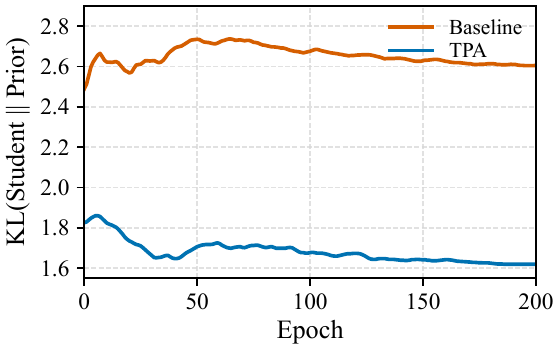}
        \caption{Student--prior KL}\label{fig:tpa-prior-kl}
    \end{subfigure}
    \caption{CSL-Daily TPA diagnostics: (a) frozen-Baichuan2-7B PPL; (b) mean student--prior KL.}
    \label{fig:tpa-fluency-kl}
\end{figure}

Across training, lower PPL accompanies lower student--prior KL with TPA (Figure~\ref{fig:tpa-fluency-kl}). Their agreement links the external fluency gain to reduced prior drift.

\noindent\textbf{Syntactic and grammatical analysis.}
Table~\ref{tab:tpa-syntax-grammar} tests whether the pattern extends to linguistic structure. Stanza parses omit punctuation and the \texttt{root} relation. Dependency JSD compares hypothesis and reference relation histograms; its $0.3$/$0.5$/$0.7$ exceedance rates capture mismatch severity, while normalized dependency edit divides relation-sequence Levenshtein distance by the longer sequence. Grammar metrics cover function-marker omissions per 100 sentences, question-particle precision, and structural-particle recall.

\begin{table}[tbp]
    \centering
    {\scriptsize
    \setlength{\tabcolsep}{2pt}
    \begin{tabular}{@{}lrrr@{}}
        \toprule
        \multicolumn{4}{@{}l}{\textit{Syntactic metrics (baseline vs. TPA)}} \\
        Metric & Base & TPA & $\Delta$ \\
        \midrule
        Mean dependency JSD $\downarrow$ & 0.4337 & \textbf{0.3955} & \deltadown{8.81\%} \\
        JSD $>0.3$ $\downarrow$ & 73.89\% & \textbf{66.07\%} & \deltadown{7.82 pp} \\
        JSD $>0.5$ $\downarrow$ & 35.12\% & \textbf{28.66\%} & \deltadown{6.46 pp} \\
        JSD $>0.7$ $\downarrow$ & 11.48\% & \textbf{7.14\%} & \deltadown{4.34 pp} \\
        Dependency edit $\downarrow$ & 0.6551 & \textbf{0.6139} & \deltadown{6.29\%} \\
        \midrule
        \multicolumn{4}{@{}l}{\textit{Grammatical metrics (baseline vs. TPA)}} \\
        Metric & Base & TPA & $\Delta$ \\
        \midrule
        Function-marker omissions $\downarrow$ & 62.1 & \textbf{56.9} & \deltadown{8.37\%} \\
        Question-particle precision $\uparrow$ & 39.6\% & \textbf{46.8\%} & \deltaup{7.2 pp} \\
        Structural-particle recall $\uparrow$ & 50.9\% & \textbf{53.7\%} & \deltaup{2.8 pp} \\
        \bottomrule
    \end{tabular}
    }
    \caption{CSL-Daily TPA linguistic diagnostics (epoch 200; omissions per 100 sentences; pp: percentage points).}
    \label{tab:tpa-syntax-grammar}
\end{table}

The distributional, sequence, and marker-level diagnostics in Table~\ref{tab:tpa-syntax-grammar} agree: TPA reduces dependency mismatch and improves particle use. Together with Figure~\ref{fig:tpa-fluency-kl}, they answer RQ1: prior preservation coincides with better-formed output in the matched comparison.

\noindent\textbf{RQ2: Does OTA improve grounding and evidence use?}
With SpaMo on CSL-Daily, Figures~\ref{fig:ota-separation-retrieval} and~\ref{fig:ota-token-lexical} compare matched OTA/no-OTA models on lexical discrimination, retrieval strata, and category/error behavior. Figure~\ref{fig:ota-occlusion} evaluates the baseline and full \method{} with matched occlusion.

\noindent\textbf{Hard-negative separation and retrieval-quality analysis.}
Figure~\ref{fig:ota-hard-negative} compares sample-level lexical margins against semantically similar hard negatives that differ in action, object, attribute, time, or quantity. For each of 348 eligible examples, $M_i=S_{\mathrm{lex}}(\widehat{Y}_i,D_i^{+})-S_{\mathrm{lex}}(\widehat{Y}_i,D_i^{-})$, where $S_{\mathrm{lex}}$ is content-token overlap and $D_i^{+}$ and $D_i^{-}$ are the visually sensitive token sets of the correct and negative references. Positive margins favor the video-specific reference. Figure~\ref{fig:ota-retrieval-quality} partitions the 1,176 test examples into quartiles of sentence-level video--reference cosine similarity (Q1 lowest; Q4 highest) and reports the share with IDF-weighted content-token recall below $50\%$. Recall is the recovered share of reference content-token IDF mass; lower low-recall rates are better.

\begin{figure}[htbp]
    \centering
    \begin{subfigure}[t]{0.5\linewidth}
        \centering
        \includegraphics[width=\linewidth]{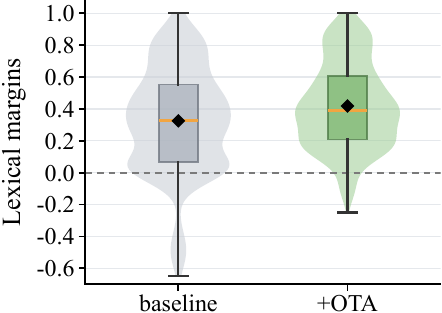}
        \caption{Hard-negative margins}\label{fig:ota-hard-negative}
    \end{subfigure}\hfill
    \begin{subfigure}[t]{0.50\linewidth}
        \centering
        \includegraphics[width=\linewidth]{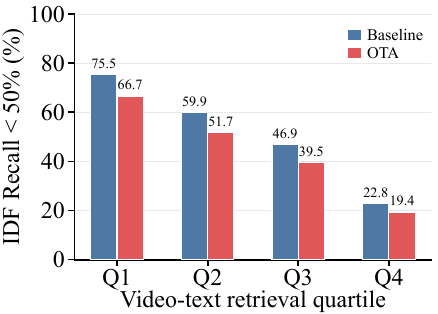}
        \caption{Low-recall rate by quartile}\label{fig:ota-retrieval-quality}
    \end{subfigure}
    \caption{OTA grounding diagnostics on CSL-Daily: (a) lexical margins against semantically similar hard negatives; (b) low-recall rates across video--text similarity quartiles.}
    \label{fig:ota-separation-retrieval}
\end{figure}

The two panels separate local grounding from coarse retrieval quality. OTA better distinguishes the correct content from matched confounders, and its low-recall advantage persists within every sentence-level similarity quartile. This advantage spans the observed similarity range.

\noindent\textbf{Token-level lexical analysis.}
At the analyzed checkpoint, Figure~\ref{fig:ota-token-category} groups reference content tokens into actions, objects, times/numbers, places, and attributes and reports IDF-weighted recall; annotations give absolute OTA--baseline differences in percentage points. Across all 1,176 test examples, Figure~\ref{fig:ota-lexical-errors} reports the percentage exhibiting omission, substitution, over-generalization, or time/number errors; annotations give relative changes from the baseline. Higher recall and lower error rates are better.

\begin{figure}[htbp]
    \centering
    \begin{subfigure}[c]{0.5\linewidth}
        \centering
        \includegraphics[width=\linewidth]{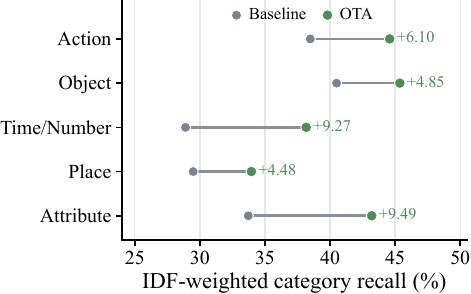}
        \caption{Category-wise IDF recall}\label{fig:ota-token-category}
    \end{subfigure}\hfill
    \begin{subfigure}[c]{0.5\linewidth}
        \centering
        \includegraphics[width=\linewidth]{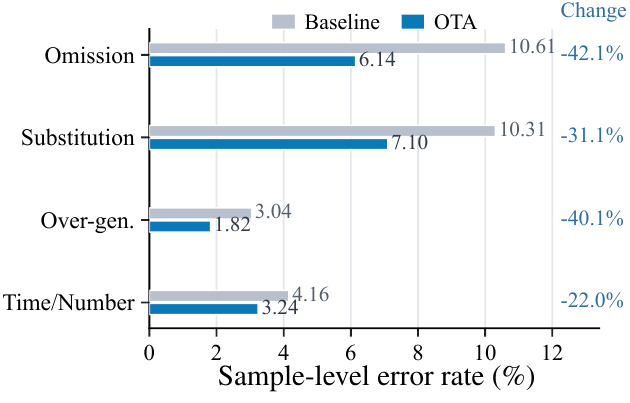}
        \caption{Sample-level error rates}\label{fig:ota-lexical-errors}
    \end{subfigure}
    \caption{CSL-Daily OTA diagnostics: (a) category-wise IDF-weighted recall; (b) sample-level lexical-error rates.}
    \label{fig:ota-token-lexical}
\end{figure}

The category and error views agree: the recall advantage spans distinct content types and coincides with fewer examples exhibiting each measured lexical error. Together, the two views support broad lexical recovery.

\noindent\textbf{OTA visual occlusion analysis.}
Figure~\ref{fig:ota-occlusion} tests whether OTA transport mass identifies visual evidence used during generation. With decoding fixed, the baseline and OTA are evaluated on the original video and with equal-length masks over random or highest-mass spans. We report absolute IDF-weighted recall and BLEU-4, not precomputed drops; larger changes indicate greater masking sensitivity.

\begin{figure}[htbp]
    \centering
    \includegraphics[width=0.96\linewidth]{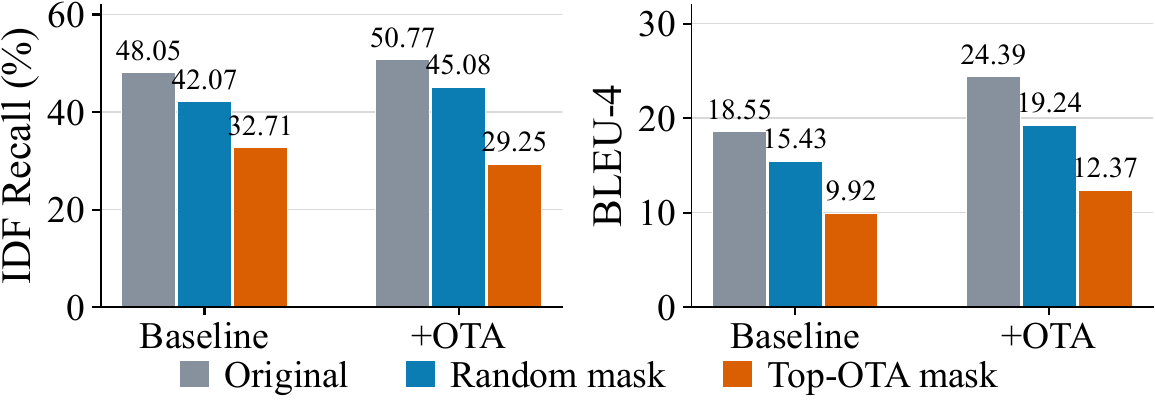}
    \caption{Absolute CSL-Daily IDF-weighted recall (left) and BLEU-4 (right) under original, random-mask, and Top-OTA-mask conditions for the baseline and OTA.}
    \label{fig:ota-occlusion}
\end{figure}

Equal-length high-transport masks disrupt content recovery and sentence-level quality more than random masks for both models. The analyses show that OTA improves fine-grained grounding and locates generation-sensitive visual evidence via transport mass.

\smallskip
\noindent\textbf{RQ3: How sensitive are TPA and OTA to their respective loss weights?}
We vary one auxiliary-loss coefficient while fixing all other training and decoding settings. Figure~\ref{fig:strength-sensitivity} reports each run's maximum corpus BLEU-4 over saved checkpoints; zero denotes the corresponding no-loss baseline.

\begin{figure}[htbp]
    \centering
    \begin{subfigure}[t]{0.5\linewidth}
        \centering
        \includegraphics[width=\linewidth]{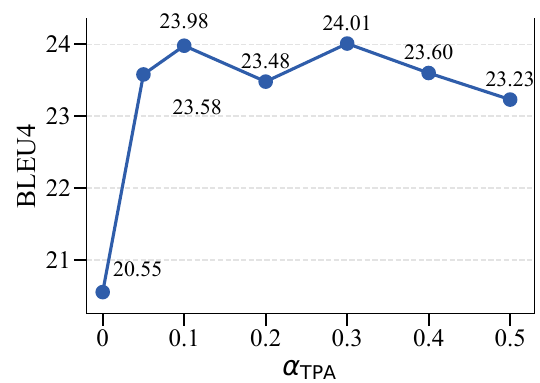}
        \caption{TPA strength}\label{fig:tpa-strength}
    \end{subfigure}\hfill
    \begin{subfigure}[t]{0.5\linewidth}
        \centering
        \includegraphics[width=\linewidth]{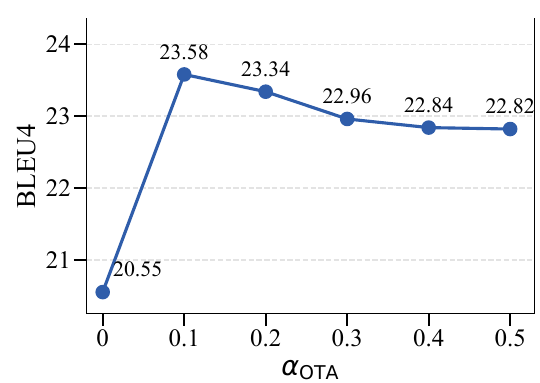}
        \caption{OTA strength}\label{fig:ota-strength}
    \end{subfigure}
    \caption{CSL-Daily best-checkpoint BLEU-4 versus (a) $\alpha_{\mathrm{TPA}}$ and (b) $\alpha_{\mathrm{OTA}}$; zero denotes no auxiliary loss.}
    \label{fig:strength-sensitivity}
\end{figure}

The sweeps distinguish robustness from insensitivity: each objective remains effective across adjacent weights, but translation quality weakens when either term grows too dominant, especially OTA. Thus, the objectives tolerate moderate retuning but should remain auxiliary to translation loss.

\smallskip
\noindent\textbf{RQ4: How do core TPA and OTA design choices affect translation quality and module behavior?}
Table~\ref{tab:design-ablation}(a) compares TPA KL directions and distillation prefixes against no TPA, pairing BLEU-4 with frozen-model PPL. Panel (b) compares sentence-level, attention-based, hard, and OTA alignment plus two OTA subsettings, pairing BLEU-4 with IDF-weighted recall.
TPA remains advantageous across the tested directions and prefixes despite a BLEU-4--PPL trade-off. Full OTA provides the strongest joint translation and recall result; removing the dustbin or transporting all tokens weakens this balance. Thus, TPA is robust across the tested implementations, whereas OTA benefits from selective partial matching that avoids unreliable correspondences.

\begin{table}[htbp]
    \centering
    \begin{minipage}[t]{0.5\linewidth}
        \centering
        \scriptsize
        \textbf{(a) TPA Design}\par\vspace{6pt}
        \begin{tabular}{@{}l@{\hspace{3pt}}r@{\hspace{3pt}}r@{}}
            \toprule
            Setting & B4 $\uparrow$ & PPL $\downarrow$ \\
            \midrule
            Baseline & 20.55 & 317.90 \\
            \addlinespace[2pt]
            \multicolumn{3}{@{}l}{\itshape Direction of KL Optimization} \\
            \hspace{0.5em}TPA w/ Forward KL & 23.37 & 265.92 \\
            \hspace{0.5em}TPA w/ Reverse KL & \textbf{23.75} & 250.84 \\
            \addlinespace[2pt]
            \multicolumn{3}{@{}l}{\itshape Context Prefix} \\
            \hspace{0.5em}TPA w/ English Context & 22.10 & \textbf{222.90} \\
            \hspace{0.5em}TPA w/ Target Context & 23.56 & 249.46 \\
            \bottomrule
        \end{tabular}
    \end{minipage}\hfill
    \begin{minipage}[t]{0.5\linewidth}
        \centering
        \scriptsize
        \textbf{(b) OTA Design}\par\vspace{2pt}
        \begin{tabular}{@{}l@{\hspace{3pt}}r@{\hspace{3pt}}r@{}}
            \toprule
            Design & B4 $\uparrow$ & IDF Recall $\uparrow$ \\
            \midrule
            \multicolumn{3}{@{}l}{\itshape Alignment Design} \\
            Sentence-level & 20.55 & 44.32 \\
            Attention alignment & 22.27 & 47.88 \\
            Hard alignment & 22.38 & 52.58 \\
            \textbf{OTA} & \textbf{24.07} & \textbf{54.20} \\
            \addlinespace[2pt]
            \multicolumn{3}{@{}l}{\itshape OTA Subsettings} \\
            \hspace{0.8em}No dustbin & 22.93 & 53.05 \\
            \hspace{0.8em}All tokens & 23.50 & 52.32 \\
            \bottomrule
        \end{tabular}
    \end{minipage}
    \caption{TPA and OTA design ablations on CSL-Daily: (a) KL direction and context prefix, with Target Context as the default; (b) alignment design, dustbin use, and token scope.}
    \label{tab:design-ablation}
    \end{table}

\FloatBarrier
\endgroup

\flushbottom

\section{Related Work}
\label{sec:related-work}

\noindent\textbf{Gloss-Free and LLM-Based Sign Language Translation.}
Gloss-based, unified, and iterative-prototype systems bridge video and text via intermediate recognition or shared representations~\cite{Camgoz_2020_CVPR,yin-read-2020-better,zhang-etal-2023-sltunet,Yao_2023_ICCV}.
Gloss-free models use visual-language pretraining, boundary-aware features, concept queries, or contrastive sign--text learning~\cite{Zhou_2023_ICCV,Yin_2023_CVPR,lin-etal-2023-gloss,jiang-etal-2024-signclip}; unsupervised and multilingual models expand supervision~\cite{guo-etal-2024-unsupervised,tan-etal-2025-multilingual}.
LLM systems condition pretrained decoders via adapters, discrete sign tokens, factorized adaptation, motion prompts, generated descriptions, or latent plans~\cite{Wong_2024_ICLR,Gong_2024_CVPR,chen-etal-2024-factorized,hwang-etal-2025-efficient,Kim_2025_ICCV,jiang-etal-2026-think}.

\noindent\textbf{Preserving Language Priors.}
Knowledge distillation transfers sequence behavior across models and multilingual translation tasks~\cite{kim-rush-2016-sequence,sun-etal-2020-knowledge-distillation}, while InstructGPT constrains policy updates through per-token KL and pretraining gradients~\cite{NEURIPS2022_b1efde53}.
Multimodal models preserve language competence via visual experts, replay, or selective cross-modal distillation~\cite{NEURIPS2024_dc06d4d2,Lin_2024_CVPR,irawan2026lingudistill}.
TPA uses a shared gold prefix and confidence-weighted token-level KL to regularize conditional translation without an inference-time prior.

\noindent\textbf{Fine-Grained Sign-Text Alignment.}
Content words carry disproportionate semantic weight in translation~\cite{chen-etal-2020-content}, motivating lexical-unit alignment rather than a single sentence target.
Prior work spans global video--text objectives~\cite{Zhou_2023_ICCV,Yin_2023_CVPR,jiang-etal-2024-signclip}, sign--word retrieval, dense contrastive separation, segment supervision, selective negatives~\cite{Cheng_2023_CVPR,NEURIPS2024_c225136c,Low_2025_ICCV,lai-etal-2026-selective}, and optimal-transport sign codebooks~\cite{Gong_2024_CVPR}.
Entropy-regularized partial transport is differentiable and robust to unmatched units, including in cross-domain alignment~\cite{NIPS2013_af21d0c9,chapel-etal-2020-partial,pmlr-v119-chen20e}.
OTA directly learns soft many-to-many grounding between continuous visual and content-token representations without an inference-time alignment module.

\section{Conclusion}

LLM-based gloss-free \SLT{} must preserve fluent generation and ground lexical choices in visual evidence.
\method{} meets both requirements through complementary training-only objectives: TPA anchors next-token predictions to a frozen language prior, while OTA uses partial optimal transport to align visual and content tokens without forcing unreliable matches.
On PHOENIX-2014T and CSL-Daily, \method{} achieves the best BLEU-4 among the compared gloss-free methods and remains effective across three language backbones.
Analyses associate TPA with lower prior drift, stronger fluency, and better grammar; OTA recovers more content tokens, reduces lexical errors, and identifies generation-sensitive visual evidence.
Thus, preserving linguistic form and fine-grained grounding jointly enable more faithful sign language translation.



\clearpage
\bibliography{aaai2027}

\end{document}